\documentclass[conference]{IEEEtran}
\IEEEoverridecommandlockouts
\usepackage{cite}
\usepackage{amsmath,amssymb,amsfonts}
\usepackage{algorithmic}
\usepackage{graphicx}
\usepackage{textcomp}
\usepackage{xcolor}

\usepackage{algorithm}

\DeclareMathOperator*{\argmin}{arg\,min}

\def\BibTeX{{\rm B\kern-.05em{\sc i\kern-.025em b}\kern-.08em
    T\kern-.1667em\lower.7ex\hbox{E}\kern-.125emX}}
\begin{document}

\title{Personalization Disentanglement for 
Federated Learning: An explainable perspective 
}

\author{\IEEEauthorblockN{1\textsuperscript{st} Peng Yan}
\IEEEauthorblockA{\textit{Faculty of Engineering and IT} \\
\textit{University of Technology Sydney}\\
Sydney, Australia \\
yanpeng9008@hotmail.com}
\and
\IEEEauthorblockN{2\textsuperscript{nd} Guodong Long}
\IEEEauthorblockA{\textit{Faculty of Engineering and IT} \\
\textit{University of Technology Sydney}\\
Sydney, Australia \\
guodong.long@uts.edu.au}
}

\maketitle

\begin{abstract}
Personalized federated learning (PFL) jointly trains a variety of local models through balancing between knowledge sharing across clients and model personalization per client. This paper addresses PFL via explicit disentangling latent representations into two parts to capture the shared knowledge and client-specific personalization, which leads to more reliable and effective PFL. The disentanglement is achieved by a novel Federated Dual Variational Autoencoder (FedDVA), which employs two encoders to infer the two types of representations. FedDVA can produce a better understanding of the trade-off between global knowledge sharing and local personalization in PFL. Moreover, it can be integrated with existing FL methods and turn them into personalized models for heterogeneous downstream tasks. Extensive experiments validate the advantages caused by disentanglement and show that models trained with disentangled representations substantially outperform those vanilla methods.
\end{abstract}


\section{Introduction}
\label{Introduction}
With increasing attention to privacy protection, federated learning (FL)~\cite{mcmahan2017communication} has recently been a hot topic in machine learning. Vanilla FL tasks require learning a global model collaboratively by clients while keeping their data decentralized and private. Various methods have been proposed under this constraint and advanced in capturing universal knowledge from data on different clients~\cite{li2018federated, kairouz2019advances, rothchild2020fetchsgd,reddi2020adaptive,DBLP:journals/corr/abs-1909-06335}. Meanwhile, samples in FL contain universal knowledge applicable all over the federation and demonstrate their host client's bias as personalized knowledge. Then personalized federated learning (PFL) is proposed to learn many local models simultaneously by leveraging universal knowledge and client biases~\cite{mansour2020three}.

\begin{figure}[ht]
\vskip 0.2in
\begin{center}
\centerline{\includegraphics[width=1.0\columnwidth]{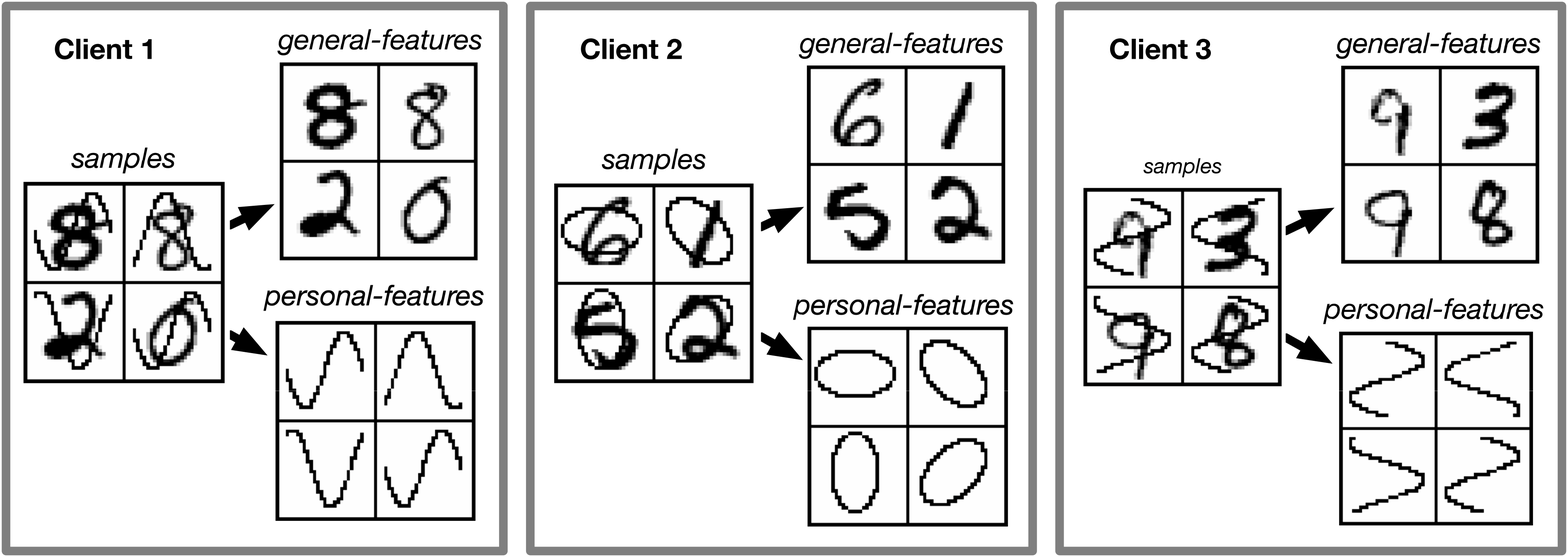}}
\caption{Suppose samples (digits) on different clients are entangled with client-related features (sinusoidal and elliptical marks). Our goal is to learn disentangled representations that each capture one type of feature.}
\label{example-disentangled-representation}
\end{center}
\vskip -0.2in
\end{figure}

In a PFL task, clients will first update local models with entangled raw sample representations, which contain both the universal and personalized knowledge, and then eliminate personalized impacts to update a global model to share universal knowledge. Most works perform these operations by designing new model architectures or adapting optimization strategies~\cite{lin2020ensemble,wang2020federated,li2021ditto,fallah2020personalized, cheng2021fine}. 
However, how to solve the PFL challenge from representation perspectives still need to be studied. One can encode a sample as two disentangled representations, each capturing one type of the above knowledge. Hence other FL algorithms for downstream tasks may learn from them separately, facilitating in extracting and sharing of universal knowledge. On the other hand, disentangled representation of personalized knowledge will help identify essential knowledge constituting a client's personality, which can support a better understanding of the locally learned models. An example of the disentangled sample representations is in Fig.\ref{example-disentangled-representation}.

To this end, we develop a federated dual variational autoencoding framework (FedDVA), where clients in the federation share two encoders inferring the above representations. The two encoders are trained collaboratively by fundamental FL algorithms like FedAvg~\cite{mcmahan2017communication}. Clients update the encoders locally by maximizing client-specific evidence lower bound (ELBO). Then a server collects local updates and aggregates them by averaging parameters. Moreover, the two encoders are cascaded and constrained by different prior knowledge so that each encoder will capture only one type of knowledge mentioned above.

We evaluate the performance of FedDVA from the perspective of disentanglement and PFL tasks. First, we explore the disentangled representations through reconstructed samples, demonstrating two different data manifolds corresponding to universal and personalized knowledge.
In addition, we train off-the-shelf FL models on the disentangled representations for classification tasks and show they will converge fast and achieve better accuracy despite various client personalities.

The main contributions of this work are summarized as follows:
\begin{itemize}
\item We propose a novel FedDVA method to solve the PFL challenge from representation perspectives. It infers disentangled representations of universal knowledge and personalized knowledge in FL. 
\item We introduce a client-specific ELBO to optimize FedDVA and analyze its capability in personalization.
\item Experiments on real-world datasets validate FedDVA's effectiveness in disentanglement and show that FL models will converge fast and achieve competitive classification performance when trained on disentangled representations.
\end{itemize}

\section{Related Work}
\textbf{Representation Disentanglement} 
Disentanglement has been extensively studied in unsupervised learning. It refers to learning a representation where a change in one dimension corresponds to a change in one factor of variation of the sample while being relatively invariant to changes in other factors \cite{bengio2013representation}. Variational autoencoder (VAE) and its variations are popular frameworks for learning disentangled representations~\cite{gregor2018temporal,chen2019isolating}. It is attractive for elegant theoretical backgrounds and high computation efficiency. However, \cite{locatello2019challenging} proved that unsupervised learning of disentangled representations is fundamentally impossible without inductive biases on both the models and the data. Then existing VAE models will fail in disentangling universal knowledge and client preferences in FL since there needs to be supervised information or ad-hoc inductive biases to distinguish between them.

\textbf{Personalized FL}
Most methods solve the PFL challenge by designing new model architecture or adapting optimization strategies. \cite{shamsian2021personalized} designed a central hypernetwork model, which is trained to generate a set of models, one model for each client. \cite{cheng2021fine} proved that steps as simple as fine-tuning the client's local data would improve the performance of a global model. \cite{fallah2020personalized} adapts the idea from MAML~\cite{finn2017model}, where clients co-learn an initialization for local models through the FedAvg. \cite{li2021ditto} proposed the Ditto, where clients share global parameters to constrain the learning process of personalized models. 

\section{Personalization Disentanglement for Federated Learning}
\begin{figure}[ht]
    \begin{center}
    \centerline{\includegraphics[width=1.0\columnwidth]{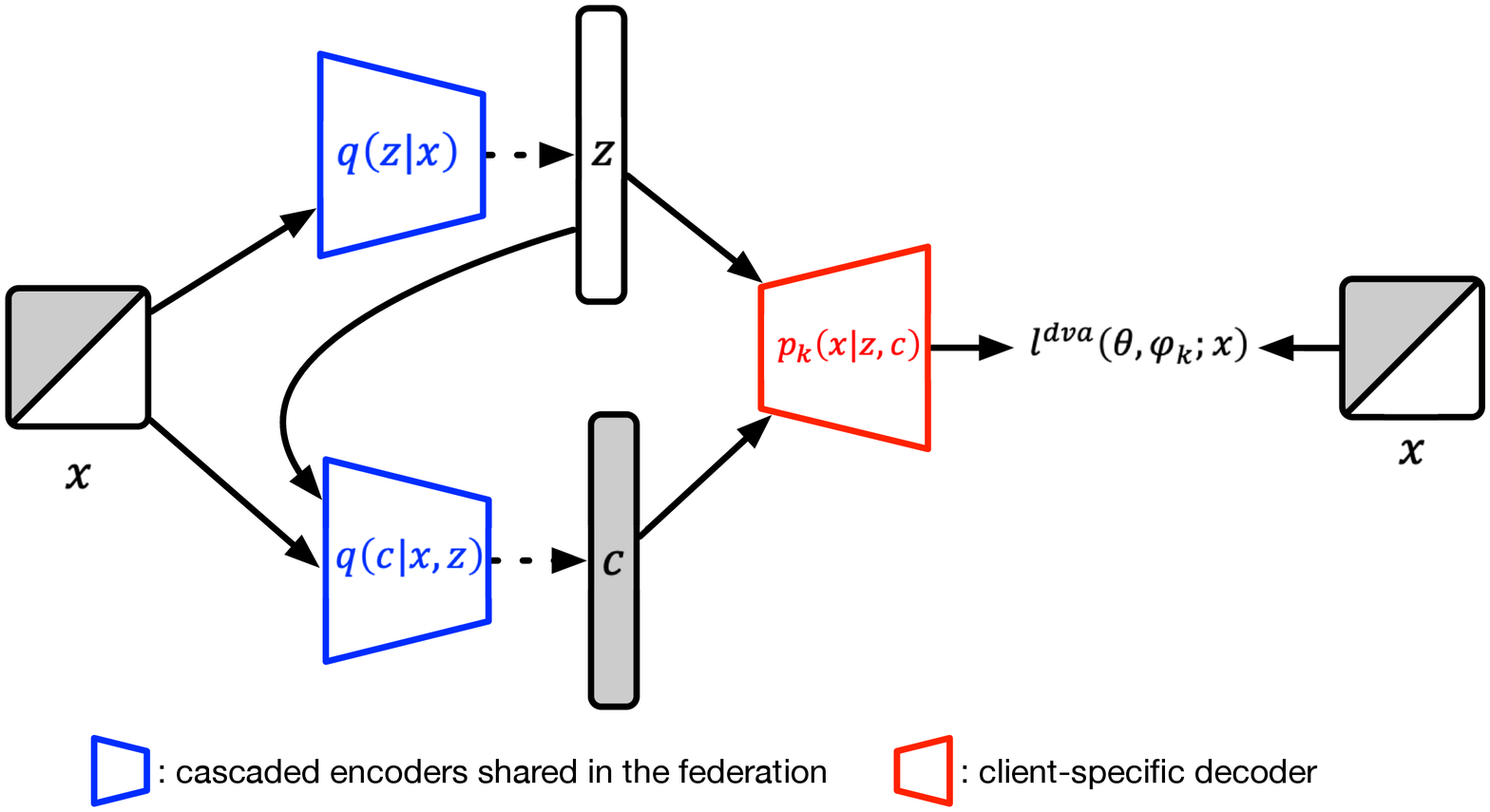}}
    \caption{The architecture of FedDVA. An encoder $f(x)$ (Blue) will first infer the posterior $q(z|x)$, and then another encoder $h(x,z)$ (red) will infer the conditional posterior $q(c|x,z)$. The decoder $g(z,c)$ (white) will try to reconstruct $x$ from $z$ and $c$.}
    \label{architecture}
    \end{center}
\end{figure}

\subsection{Background and Notation}
\textbf{Federated Learning} In a federated learning system with $K$ clients, each client is indexed by $k$ with its local data denoted as $\mathcal{D}_k$. The objective of a federated learning task is to find the optimal parameters $\theta^{*}$ sharing with all clients, which is formulated as
\begin{equation}
    \begin{aligned}
    \theta^{*}=\argmin_{\theta}\sum_{k=1}^{K} w_k \mathcal{L}_{k}(\theta;\mathcal{D}_{k})
    \end{aligned}
    \label{eq-loss-fedavg}
\end{equation}
where $\mathcal{L}_{k}(\theta;\mathcal{D}_{k})=\sum_{x\in\mathcal{D}_{k}}\ell(\theta;x)$ is the learning objective on the $k$-th client, $w_k=|\mathcal{D}_k|/\sum_{k=1}^{K} |\mathcal{D}_k|$ is an importance weight for the client, and $\ell(\theta;x)$ is a loss function

\textbf{VAE Framework}
VAE assumes any sample $x$ corresponds to a latent representation $z$ from the prior $p(z)=\mathcal{N}(z;0,I)$. It learns an encoder to infer the variational posterior $q_{\theta}(z|x)$ and a decoder to reconstruct the sample $x$ from $z\sim q_{\theta}(z|x)$. In general, the encoder is a neural network whose outputs are the mean and covariance of the variational posterior $q_{\theta}(z|x)$, that is, $q_{\theta}(z|x)=\mathcal{N}(z;\mu(x),\Sigma(x))$. The covariance matrix $\Sigma$ is assumed to be diagonal for computation simplicity. The decoder is another neural network generating $x$ by maximizing the log-likelihood $\log p_{\varphi}(x|z)$. $\theta$ and $\varphi$ are learnable parameters of the encoder and the decoder. They can be optimized by maximizing the below ELBO
\begin{equation}
    \begin{aligned}
    EB(x)=\mathbb{E}_{q_{\theta}(z|x)}[\log p_{\varphi}(x|z)]-D_{KL}(q_{\theta}(z|x)||p(z))
    \end{aligned}
    \label{eq-vae-elbo}
\end{equation}
The first term on the RHS of Eq.\ref{eq-vae-elbo} measures the reconstruction performance of latent representation $z$, and the second term measures the $KL$-divergence between the posterior $q_{\theta}(z|x)$ and the prior $p(z)$. Gradient-based optimization methods apply with the reparameterization trick \cite{kingma2013auto}. In the rest of the paper, we will use $\ell(\theta,\varphi;x)$ to denote the negative ELBO and omit the subscripts $\theta$ and $\varphi$ for notational simplicity, i.e.,
\begin{equation}
    \begin{aligned}
    \ell(\theta,\varphi;x)=-\mathbb{E}_{q(z|x)}[\log p(x|z)]+D_{KL}(q(z|x)||p(z))
    \end{aligned}
    \label{eq-vae-free}
\end{equation}

\subsection{Methodology}
\textbf{Problem Formulations} The goal of our method is to learn disentangled sample representations for universal knowledge and personalized knowledge. We denote them as $z$ and $c$. Since $z$ is irrelevant to clients, we assume samples on each client have the same prior distribution $p(z)=\mathcal{N}(z;0, I)$. Meanwhile, since samples in FL are private and distributed, the prior distribution of $c$ is unknown and varies among clients. We denote it as $p_k(c)$, where $k$ is the index of the $k$-th client. But we can not make assumptions about the $p_{k}(c)$ as we can not ensure that the relationship between the assumed one is consistent with the relationship between clients. For example, we can only allocate the same $p_k(c)$ to two clients after disclosing that they have similar personalities. Alternatively, we assume the prior distribution of the data in the federation is standard Gaussian, or equivalently, the mixture distribution of local priors $q(c)=\sum_{k=1}^{K}w_{k}p_{k}(c)$ is $\mathcal{N}(c;0,I)$. 

\textbf{Dual Encoders}
As illustrated in Fig.\ref{architecture}, the proposed FedDVA learns the above representations through two encoders. For any sample $x$, an encoder first infers variational posterior $q(z|x)=\mathcal{N}(z;\mu(x),\Sigma(x))$ for the universal knowledge $z$. Then another encoder infers variational posterior $q(c|x,z)=\mathcal{N}(c;\hat{\mu}(x,z),\hat{\Sigma}(x,z))$, conditioned on both the sample $x$ and the representation $z$, for the impacts of personalities. In addition, a client-specific local decoder will evaluate the reconstruction performance of representations $z$ and $c$. It is implemented by a neural network maximizing the client-specific log-likelihood $\log p_{k}(x|z,c)$. The negative ELBO optimizing FedDVA is in Eq.\ref{eq-feddva-loss}
\begin{equation}
    \begin{aligned}
    \ell^{cva}(\theta,\varphi_{k};x)=-\mathbb{E}_{q(z|x)}[\mathbb{E}_{q(c|x,z)}[\log p_{k}(x|z,c)]\\+\beta\mathcal{R}_{c}(q(c|x,z))]+\alpha\mathcal{R}_{z}(q(z|x))
    \end{aligned}
    \label{eq-feddva-loss}
\end{equation}
$\theta$ in the Eq.\ref{eq-feddva-loss} denotes parameters of the shared encoders, $\varphi_k$ denotes the parameters of the local decoder specific to the $k$-th client, $\mathcal{R}_{z}(q(z|x))$ and $\mathcal{R}_{c}(q(c|x,z))$ denote the regularizers for the posterior $q(z|x)$ and $q(c|x,z)$, $\alpha$ and $\beta$ are their importance weights.

Similar to traditional VAE models, the posterior $q(z|x)$ is regularized by $D_{KL}(q(z|x)||p(z))$, which enforces the distribution of the representation $z$ to approximate the standard Gaussian distribution. But it would be challenging to regularize the representation $c$ without prior knowledge about the distribution $p_{k}(c)$. FedDVA handles the problem by a slack regularizer $D_{KL}(q(c|x,z)||q(c))$ combing with a constraint that
\begin{equation}
    \begin{aligned}
    D_{KL}(q(c|x,z)||q(c))-D_{KL}(q(c|x,z)||\bar{p}_{k}(c))\ge\xi_{k}
    \end{aligned}
    \label{ieq-reg-c}
\end{equation}
where $\bar{p}_{k}(c)=\frac{1}{|\mathcal{D}_{k}|}\sum_{x\in\mathcal{D}_{k}}q(c|x,z)$ is the mixture distribution of $q(c|x,z)$ of samples on the $k$-th client, and $\xi_{k}>0$ is a hyperparameter. Intuitively, $\bar{p}_{k}(c)$ is an estimator of $p_{k}(c)$, and the Ineuqation.\ref{ieq-reg-c} requires $q(c|x,z)$ to be at least $\xi_{k}$ closer to $\bar{p}_{k}(c)$ than $q(c)$. We will discuss it in Sec.\ref{sec-theoretical-analysis} and show it helps the representation $c$ to capture client personalities. Combining the $KL$-divergence and the constraint in Inequation.\ref{ieq-reg-c}, regularizers $\mathcal{R}_{z}(q(z|x))$ and $\mathcal{R}_{c}(q(c|x,z))$ of Eq.\ref{eq-feddva-loss} are 
\begin{equation}
    \begin{aligned}
    &\mathcal{R}_{z}(q(z|x))=D_{KL}(q(z|x)||p(z))
    \label{eq-reg-z}
    \end{aligned}
\end{equation}
\begin{equation}
    \begin{aligned}
    \mathcal{R}_{c}(q(c|x,z))=\max(&\xi_{k}+D_{KL}(q(c|x,z)||\bar{p}_{k}(c))\\
    &, D_{KL}(q(c|x,z)||q(c)))
    \label{eq-reg-c}
    \end{aligned}
\end{equation}
They can be computed and differentiated without estimation (see Appendix B). Accordingly, the learning problem of FedDVA can be solved by gradient-based methods.

\textbf{Optimization}
To learn the encoders collaboratively, we formulate the learning objective of FedDVA as follows:
\begin{equation}
    \begin{aligned}
    \argmin_{\theta,\varphi_{1}...\varphi_{K}}\sum_{k=1}^{K}w_{k}\mathcal{L}_{k}(\theta,\varphi_k;\mathcal{D}_{k})
    \end{aligned}
    \label{eq-fedcva-objective}
\end{equation}
where $\mathcal{L}_{k}(\theta,\varphi_{k};\mathcal{D}_{k})=\sum_{x\in\mathcal{D}_{k}}\ell^{cva}(\theta,\varphi_k;x)$. Then gradient steps optimizing Eq.\ref{eq-fedcva-objective} consist of the following two parts
\begin{equation}
    \begin{aligned}
    \varphi_{k}'=\varphi_{k}-\eta\nabla_{\varphi_k}\mathcal{L}_{k}(\theta,\varphi_k;\mathcal{D}_{k}), 1\le k\le K
    \end{aligned}
    \label{eq-gradient-local-decoder}
\end{equation}
\begin{equation}
    \begin{aligned}
    \theta'=\theta-\lambda\sum_{k=1}^{K}w_{k}\nabla_{\theta}\mathcal{L}_{k}(\theta,\varphi_{k};\mathcal{D}_{k})
    \end{aligned}
    \label{eq-gradient-local-encoder}
\end{equation}
where $\eta$ and $\lambda$ are their learning rates. Eq.\ref{eq-gradient-local-decoder} updates the client-specific decoders and is processed by each client independently. Eq.\ref{eq-gradient-local-encoder} updates the shared encoders shared in the federation. Most FL algorithms like FedAvg can optimize it. Concretely, $\theta'=\sum_{k=1}^{K}w_{k}\theta'_{k}$, where
\begin{equation}
    \begin{aligned}
     \theta'_{k}=\theta_{k}-\lambda\nabla_{\theta}\mathcal{L}_{k}(\theta_{k},\varphi_{k};\mathcal{D}_{k}), 1\le k\le K
    \end{aligned}
    \label{eq-gradient-local-encoder-fedavg}
\end{equation}
, and Eq.\ref{eq-gradient-local-encoder-fedavg} is performed by each client independently.
But it is worth noting that the optimization steps of Eq.\ref{eq-gradient-local-decoder} and Eq.\ref{eq-gradient-local-encoder-fedavg} are asynchronous. As only a subset of clients will participate in the optimization process in each communication round \cite{mcmahan2017communication}, client-specific decoders may not coincide with the shared encoders. A client needs to update $\varphi_k$ first and later the $\theta$. Complete pseudo-codes of the optimization process are in Algorithm.\ref{alg:FedDVA}.
\begin{algorithm}[tb]
   \caption{FedDVA}
   \label{alg:FedDVA}
   \hspace*{0.02 in} {\bfseries Input:} $m$: number of clients sampled each round; $b$: batch size; $\lambda$ and $\eta$: learning rates; $\xi_{k}$: the constraint threshold in Inequation (\ref{ieq-reg-c}).\\
   \hspace*{0.02in} {\bf Server executes:}
   \begin{algorithmic}
   \STATE Initialize $\theta^{(1)}\leftarrow\theta$
   \FOR{each round $r=1,2,...$}
   \STATE Sample a set of $m$ clients $\mathbb{C}$
   \FOR {each client $k\in \mathbb{C}$ \textbf{parallel}}
   \STATE $\theta_k^{(r+1)}\leftarrow$\textbf{ClientUpdate}$(k, \theta^{(r)})$
   \ENDFOR
   \STATE $\theta^{(r+1)}\leftarrow \sum_{k=1}^{K}w_{k}\theta_{k}^{(r+1)}$
   \ENDFOR
   \end{algorithmic}
   \hspace*{0.02in} {\bf ClientUpdate$(k, \theta)$:}
   \begin{algorithmic}
   \STATE Initialize $\theta_{k}\leftarrow\theta$, $\varphi_{k}\leftarrow\varphi$
   \FOR{batch $\mathcal{B}\subset\mathcal{D}_{k}$}
   \STATE update $\varphi_{k}$ by Eq.\ref{eq-gradient-local-decoder}
   \ENDFOR
   \STATE $\varphi\leftarrow \varphi_{k}$
   \FOR{batch $\mathcal{B}\subset\mathcal{D}_{k}$}
   \STATE update $\theta_{k}$ by Eq.\ref{eq-gradient-local-encoder-fedavg}
    \ENDFOR
    \STATE return $\theta_{k}$
\end{algorithmic}
\end{algorithm}

\section{Theoretical Analysis}
\label{sec-theoretical-analysis}
In this section, we discuss the ELBO corresponding to Eq.\ref{eq-feddva-loss} and show that it has the capability to capture client personalities.

From the perspective of variational inference, the optimal posteriors $q(z|x)$ and $q(c|x,z)$ are the ones maximizing the following EBLOs jointly
\begin{equation}
    \begin{aligned}
    ELBO_{z}(x,k)=&\mathbb{E}_{q(z|x)}[\log p_{k}(x|z)]\\&-D_{KL}(q(z|x)||p(z))
    \end{aligned}
    \label{eq-elbo-z}
\end{equation}
\begin{equation}
    \begin{aligned}
    ELBO_{c}(x,z,k)=&\mathbb{E}_{q(c|x,z)}[\log p(x|z,c)]\\&-D_{KL}(q(c|x,z)||p_{k}(c))
    \end{aligned}
    \label{eq-elbo-c}
\end{equation}
where the subscript $k$ means the distribution is specific to the $k$-th client. Ideally, $\log p(x|z,c)$ is a client irrelevant log-likelihood modeling the sample generating process, that is, $p_{k}(x)=\iint p(x|z,c)p(z)p_{k}(c)\text{d}z\text{d}c$ (Details of the derivation are given in Appendix A.1 and A.2). But Eq.\ref{eq-elbo-c} is hard to compute in practice. Besides the unknown prior knowledge $p_{k}(c)$, the client irrelevant log-likelihood $\log p(x|z,c)$ is unavailable in FL. For example, sharing $\log p(x|z,c)$ in the federation risks privacy leakage as it has the capability to generate samples.

As an alternative, FedDVA optimizes the posterior $q(c|x,z)$ by maximizing the ELBO in Eq.\ref{eq-elbo-c-alternative}
\begin{equation}
    \begin{aligned}
    ELBO'_{c}(x,z,k)=&\mathbb{E}_{q(c|x,z)}[\log p_{k}(x|z,c)]\\&-D_{KL}(q(c|x,z)||q(c))
    \end{aligned}
    \label{eq-elbo-c-alternative}
\end{equation}
which is equivalent to Eq.\ref{eq-elbo-c}, except for that the slack regularizer $D_{KL}(q(c|x,z)||q(c))$ degenerates the capability of capturing differences between clients. Specifically, the overall $KL$-divergence between $q(c|x,z)$ and $q(c)$ of samples on the same client is
\begin{equation}
    \begin{aligned}
    \mathbb{E}_{p_{k}(x)}[\mathbb{E}_{q(z|x)}[-H(q(c|x,z))]]+H(\bar{p}_{k}(c),q(c))
    \end{aligned}
\end{equation}
which requires the distribution of representation $c$ to be close to $q(c)$ wherever the samples are. Inequation.\ref{ieq-reg-c} helps resolve the problem by introducing an inductive bias that the posterior $q(c|x,z)$ of samples on the same client is closer to $p_{k}(c)$ than to $q(c)$, with which $D_{KL}(\bar{p}_{k}(c)||q(c))\ge\xi_{k}$ holds (Details of the derivation are in Appendix A.3). Finally, replacing $p_{k}(x|z)$ in Eq.\ref{eq-elbo-z} with Eq.\ref{eq-elbo-c-alternative}, we have the loss function described in Eq.\ref{eq-feddva-loss}, and the hyperparameter $\xi_{k}$ helps determinate the degree of 'penalization' representation $c$ captured. The larger the $\xi_{k}$ is, the more personalized representation $c$ is learned.


\section{Experiment}
In this section, we verify FedDVA's disentanglement effectiveness by demonstrating manifolds of the sample reconstructed from the disentangled representations $z$ and $c$. Then, we evaluate classification performance based on representations from FedDVA. Details and full results are provided in Appendix.\ref{AppendixExperiments} and codes are also available on the Github\footnote{https://github.com/pysleepy/FedDVA}.

\subsection{Personalization Disentanglement}
\label{Personalization Disentanglement}
We empirically study FedDVA's disentanglement capability on real-world data sets with different personalization settings.

\begin{figure}[ht]
    \begin{center}
    \centerline{\includegraphics[width=.95\columnwidth]{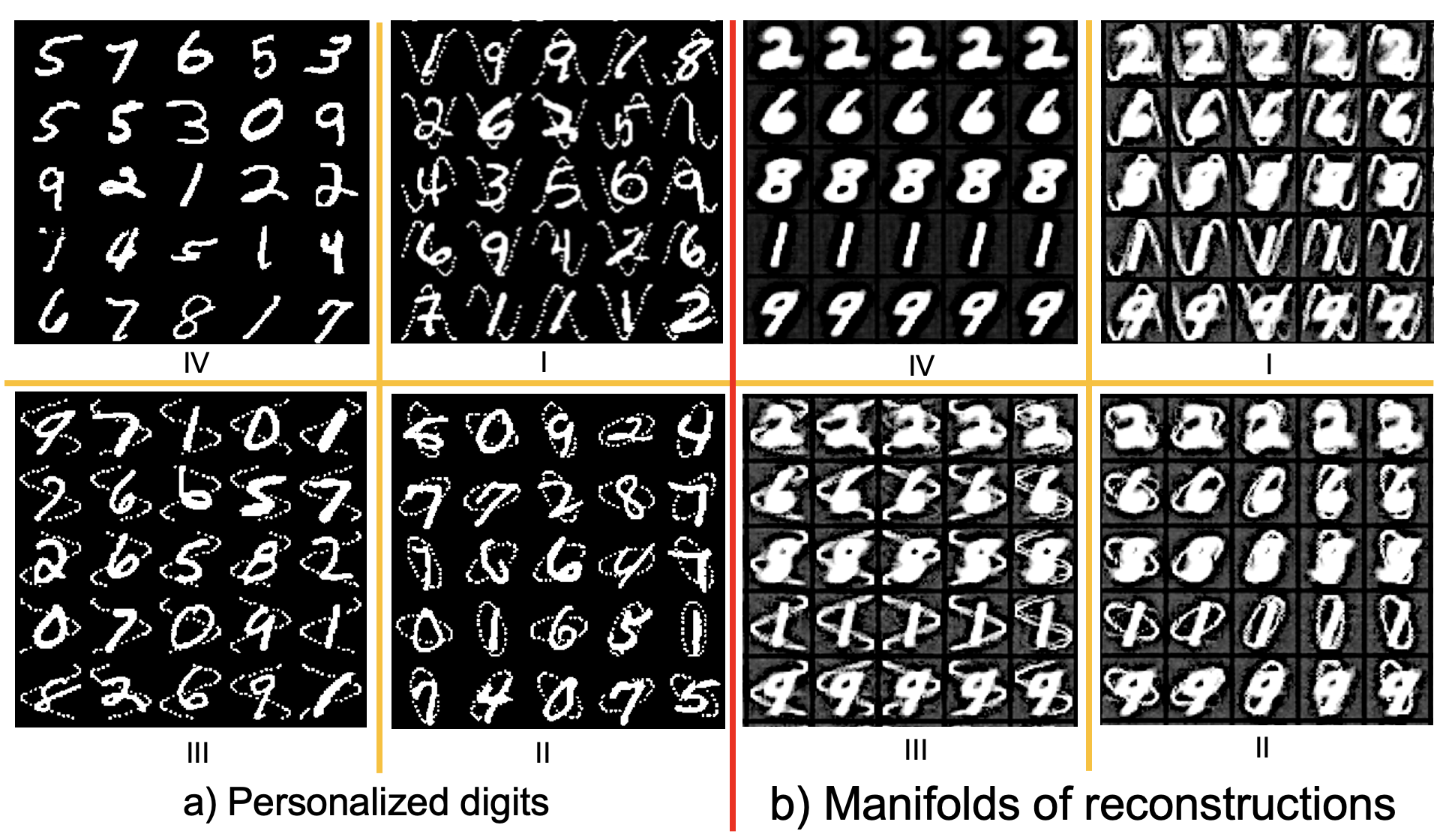}}
    \caption{a) Digits in each quadrant denote the samples from the same client. They are synthesized with client-specific marks, i.e., horizontal sinusoids, ellipses, vertical sinusoids and plain digits. b) Reconstructions will vary along with the changes in general representation $z$ (rows) while invariant to the personalized representation $c$ (columns). Vice versa for client-specific marks.}
    \label{MNISTExample}
    \end{center}
\end{figure}

\textbf{MNIST}\footnote{https://yann.lecun.com/exdb/mnist/} is a benchmark dataset of handwritten digits with 60,000 training images and 10,000 testing images. We uniformly allocate them to a set of clients and synthesize them with client-specific marks. An example is in Fig.\ref{MNISTExample}(a).

We train the FedDVA model by Algorithm.\ref{alg:FedDVA} and visualize manifolds of data reconstructed from the learned representations. Fig.\ref{MNISTExample}(b) shows that representations $z$ and $c$ are disentangled. Digits in the images will vary along with the changes in general representation $z$ (rows) while invariant to the personalized presentation $c$ (columns). Similarly, client-specific marks will vary along with the changes in $c$ and remain unchanged when $z$ changes.

\begin{figure}[ht]
    \begin{center}
    \centerline{\includegraphics[width=.95\columnwidth]{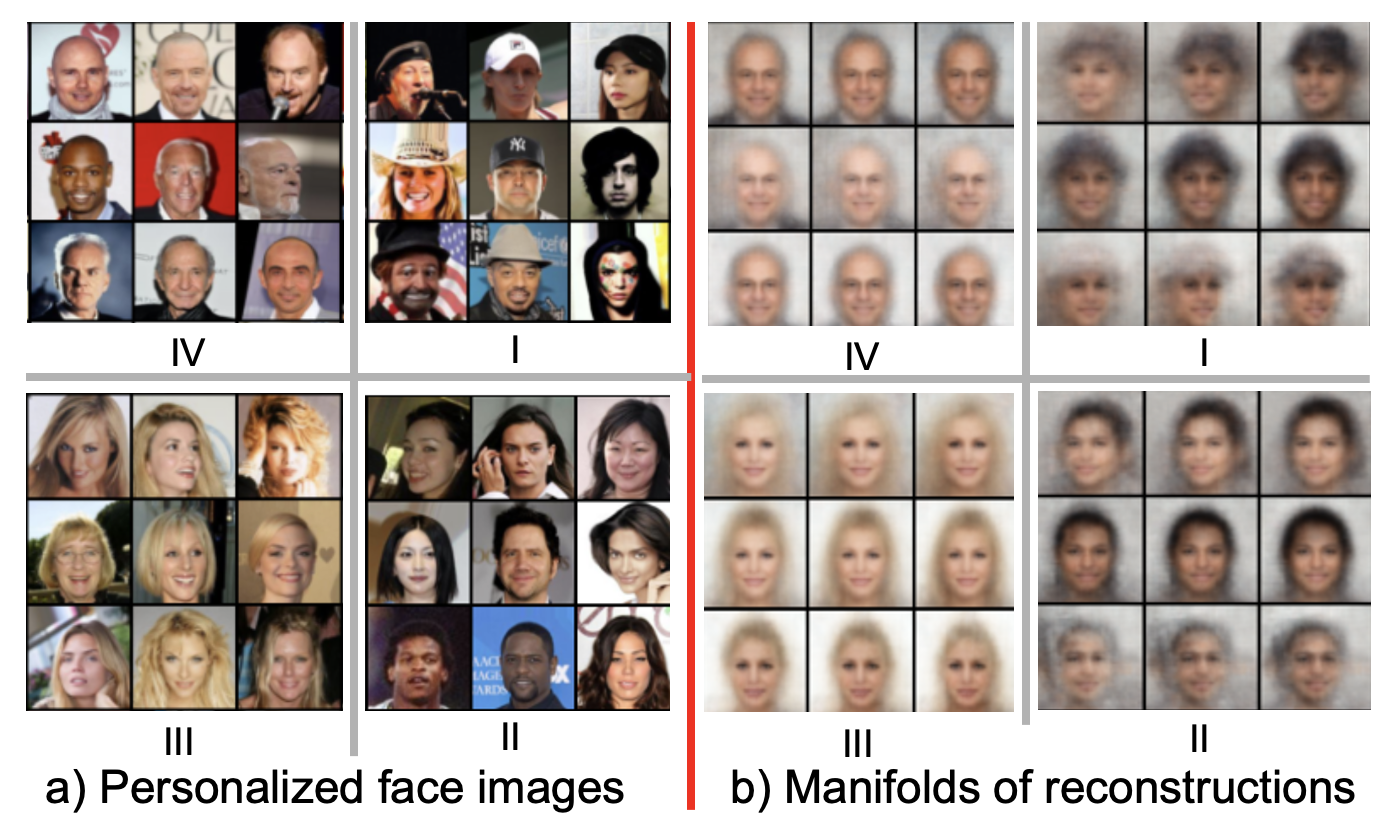}}
    \caption{a) The faces in each quadrant correspond to the data from the same client. They share similar attributes regarding their hairstyles, i.e., wearing hats, black hair, blond, and bald. b) Images generated from the same general representation $z$ (rows) will have similar faces and vary in personalized attributes regarding personalized representation $c$ (columns).}
    \label{CelebAExample}
    \end{center}
\end{figure}

\textbf{CelebA}\footnote{https://mmlab.ie.cuhk.edu.hk/projects/CelebA.html} is a large-scale face dataset containing 202,599 face images of celebrities. We allocate them to clients according to face attributes so that images on the same client will demonstrate a bias towards some attributes, e.g., hairstyles. Examples of personalized face images are shown in Fig.\ref{CelebAExample}(a).

We can find that major face attributes and hairstyles are disentangled. Faces generated from the same general representation $z$ are similar and will vary in hairstyles when personalized representation $c$ changes. Meanwhile, other significant attributes like background colors will also vary along with $z$, while miscellaneous attributes like face angles are implied as personalized knowledge in $c$.

\begin{figure}[ht]
    \begin{center}
    \centerline{\includegraphics[width=.95\columnwidth]{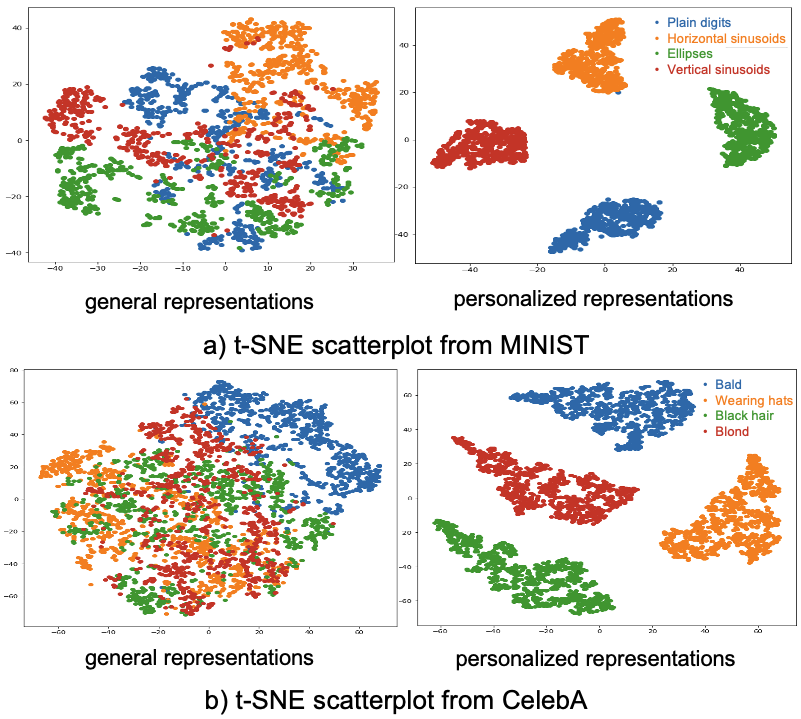}}
    \caption{Representation distributions regarding $z$ (Left) and $c$ (Right). Each dot denotes a sample's latent representation, and the color denotes the client it is sampled from. It can be found that distributions of the general representation $z$ are client irrelevant, and the counterparts of the personalized representation $c$ are clustered regarding their clients}
    \label{Scatterplot}
    \end{center}
\end{figure}

In addition, we also visualize the distribution of the representations learned by FedDVA. We embed them into the 2-dimension space by the t-SNE and visualize them by scatter plots (Fig.\ref{Scatterplot}). It can be found that distributions of the general representation $z$ (left) are mixed and client-irrelevant. The counterparts of the personalized representation $c$ (right) are clustered regarding their clients.


\subsection{Personalized Classification}
\begin{figure}[ht]
    \begin{center}
    \centerline{\includegraphics[width=.95\columnwidth]{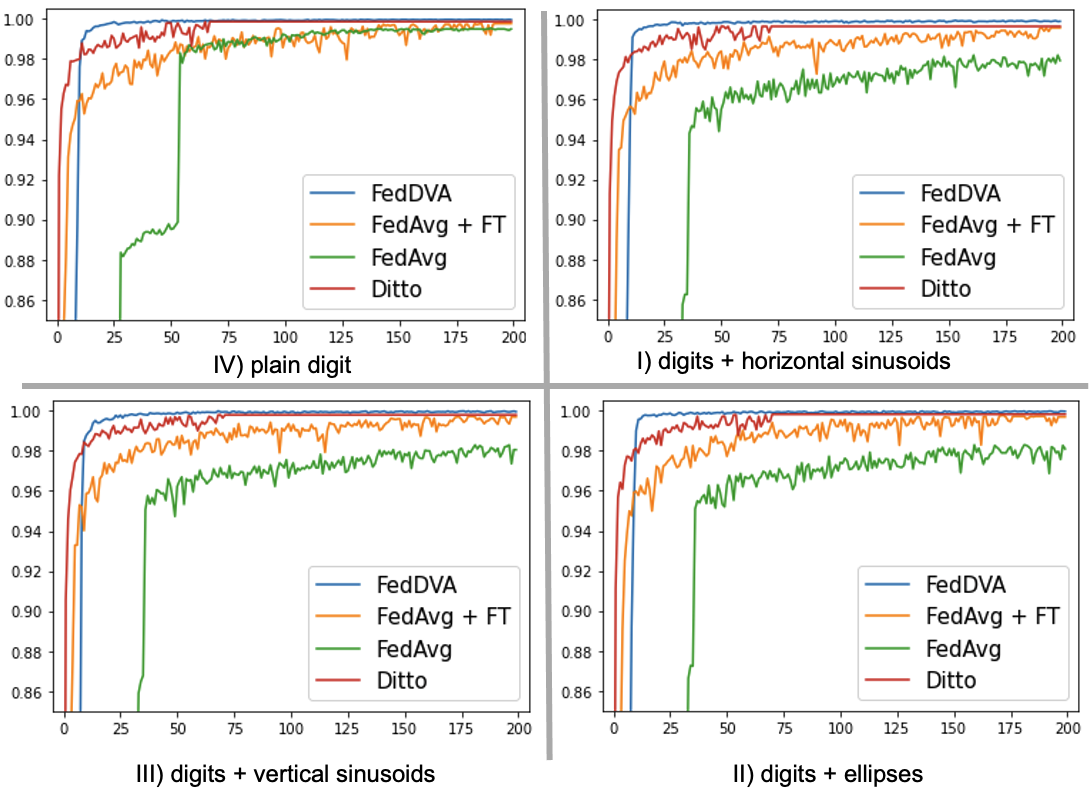}}
    \caption{Accuracy on clients where digits are synthesized with marks. FedDVA (blue) will achieve better accuracy (vertical) at the same communication round (horizontal).}
    \label{heterogeneous-x-mnist-accuracy}
    \end{center}
\end{figure}

\begin{figure}[ht]
    \begin{center}
    \centerline{\includegraphics[width=.95\columnwidth]{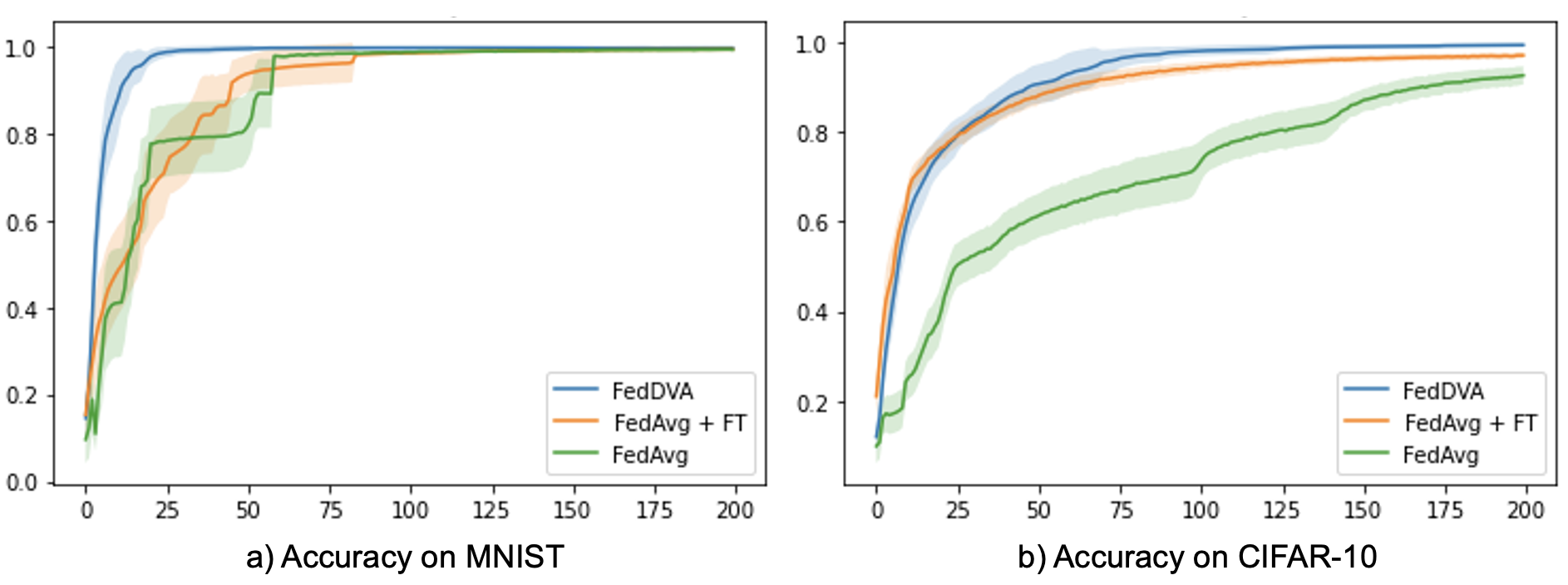}}
    \caption{Accuracy on MNIST and CIFAR0-10. FedDVA will achieve higher accuracy each communication round and have less variance regarding the accuracy among clients (shadows).}
    \label{heterogeneous-y-mnist-cifar-10-accuracy}
    \end{center}
\end{figure}

We evaluate the classification performance of representations learned by FedDVA. We tune the dual encoders along with a classification head and compare their performance with vanilla FL algorithms FedAvg~\cite{mcmahan2017communication}, FedAvg+Fine Tuning~\cite{cheng2021fine} and DITTO~\cite{li2021ditto}. Two personalization settings are applied. 1) heterogeneous inputs: digits on clients are synthesized with client-specific marks as in Fig.\ref{MNISTExample}(a); 2) heterogeneous outputs: we allocate samples to clients so that they will vary in label distributions. Two benchmark datasets, MNIST and CIFAR-10\footnote{https://www.cs.toronto.edu/~kriz/cifar.html}, are applied. Results in Fig.\ref{heterogeneous-x-mnist-accuracy} and Fig.\ref{heterogeneous-y-mnist-cifar-10-accuracy} show that a model based on the disentangled representations will converge fast and achieve competitive performance to those vanilla FL methods.

\section{Conclusion}
In conclusion, this paper proposes a novel FedDVA method to disentangle general and personalized representations for PFL. Empirical studies validate FedDVA's disentanglement capability and show that disentangled representations will improve convergence and classification performance.



\bibliographystyle{IEEEbib}
\bibliography{icme2023}

\onecolumn
\newpage
\appendix
\section{Evidence Lower Bounds}
\subsection{ELBO optimizing $q(z|x)$}
Suppose $p(z|x)$ is the true posterior of $z$, $q(z|x)$ is the variational posterior approximating $p(z|x)$ and samples on the same client are independent and identical distributed (iid.), then the learning task on the $k$-th client is to minimize $D_{KL}(q(z|x)||p(z|x))$, which is
\begin{equation}
    \begin{aligned}
    D_{KL}(q(z|x)||p(z|x))=&\int q(z|x) \log \frac{q(z|x)}{p(z|x)} \text{d}z\\
    =&\int q(z|x) \log \frac{q(z|x)p_{k}(x)}{p(z|x)p_{k}(x)} \text{d}z\\
    =&\log p_{k}(x)+\int q(z|x)\log \frac{q(z|x)}{p_{k}(x|z)p(z)}\text{d}z\\
    =&\log p_{k}(x)-\mathbb{E}_{q(z|x)}[\log p_{k}(x|z)]+D_{KL}(q(z|x)||p(z))
    \end{aligned}
\end{equation}
or equivalently,
\begin{equation}
    \begin{aligned}
    \log p_{k}(x)\ge ELBO_{z}(x,k)=\mathbb{E}_{q(z|x)}[\log p_{k}(x|z)]-D_{KL}(q(z|x)||p(z))
    \end{aligned}
\end{equation}
\subsection{ELBO optimizing $q(z|x,c)$}
Suppose $p(c|x,z)$ is the true posterior of $c$, $q(c|x,z)$ is the variational posterior approximating $p(c|x,z)$ and samples on the same client are iid., then the learning task on the $k$-th client is to minimize $D_{KL}(q(c|x,z)||p(c|x,z))$, which is
\begin{equation}
    \begin{aligned}
    D_{KL}(q(c|x,z)||p(c|x,z))=&\int q(c|x,z) \log\frac{q(c|x,z)}{p(c|x,z)}\text{d}c\\
    =&\int q(c|x,z) \log \frac{q(c|x,z)p_{k}(x|z)}{p(c|x,z)p_{k}(x|z)}\text{d}c\\
    =&\log p_{k}(x|z) + \int q(c|x,z)\log\frac{q(c|x,z)}{p_{k}(x,c|z)}\text{d}c\\
    =&\log p_{k}(x|z) - \mathbb{E}_{q(c|x,z)}[\log p_{k}(x,c|z)] - H(q(c|x,z))
    \end{aligned}
\end{equation}
Ideally, there is a client-irrelevant likelihood $p(x|z,c)$ modeling the sample generating process, that is $p_{k}(x)=\iint p(x|z,c)p(z)p_{k}(c)\text{d}z\text{d}c$, where the personality of a client lies on $p_{k}(c)$. Then we have
\begin{equation}
    \begin{aligned}
    \log p_{k}(x)\ge ELBO_{c}(x,z,k)=\mathbb{E}_{q(c|x,z)}[\log p(x|z,c)]-D_{KL}(q(c|x,z)||p_{k}(c))
    \end{aligned}
\end{equation}
which is equivalent to
\begin{equation}
    \begin{aligned}
    \log p_{k}(x)\ge ELBO'_{c}(x,z,k)=\mathbb{E}_{q(c|x,z)}[\log p_{k}(x|z,c)]-D_{KL}(q(c|x,z)||q(c))
    \end{aligned}
\end{equation}
\subsection{Difference between $D_{KL}(q(c|x,z)||q(c))$ and $D_{KL}(q(c|x,z)||\bar{p}_{k}(c))$}
For samples on the same client and suppose they are independent and identical distributed, according to Eq.\ref{ieq-reg-c}, there is
\begin{equation}
    \begin{aligned}
    &\mathbb{E}_{p_{k}(x)}[\mathbb{E}_{q(z|x)}[D_{KL}(q(c|x,z)||q(c))-D_{KL}(q(c|x,z)||\bar{p}_{k}(c))]]\\
    =&\mathbb{E}_{p_{k}(x)}[\mathbb{E}_{q(z|x)}[\mathbb{E}_{q(c|x,z)}[\log \bar{p}_{k}(c)-\log q(c)]]]\\
    =&\mathbb{E}_{\bar{p}_{k}(c)}[\log \bar{p}_{k}(c)-\log q(c)]\\
    =&D_{KL}(\bar{p}_{k}(c)||q(c))\\
    \ge&\xi_{k}
    \end{aligned}
\end{equation}

\section{Computation of the $KL$-Divergence}
\subsection{$KL$-Divergence between two Gaussian distributions}
For the $i$-th sample $x_{i}$ and $j$-th sample $x_{j}$, variational posteriors inferring representation $c$ are $q(c|x_i, z_i)=\mathcal{N}(c;\mu_i,\Sigma_i)$ and $q(c|x_j, z_j)=\mathcal{N}(c;\mu_j,\Sigma_j)$, where $c$ is a $d$-dimensional vector and convariance matrices of $\Sigma_i$ and $\Sigma_j$ are diagonal. Then we have
\begin{equation}
    \begin{aligned}
    \int q(c|x_i,z_i)\log q(c|x_i,z_i)\text{d}c=&\int \mathcal{N}(c;\mu_i,\Sigma_i) \log \mathcal{N}(c;\mu_i,\Sigma_i)\text{d}c\\
    =&-\frac{1}{2}(d\log(2\pi)+\log|\Sigma_i|+d)
    \end{aligned}
    \label{computation-hqc}
\end{equation}
and
\begin{equation}
    \begin{aligned}
    \int q(c|x_i,z_i)\log q(c|x_j,z_j)\text{d}c=&\int \mathcal{N}(c;\mu_i,\Sigma_i)\log \mathcal{N}(c;\mu_j,\Sigma_j)\text{d}c\\
    =&-\frac{1}{2}(d\log(2\pi)+\log|\Sigma_j|+Tr(\Sigma_j^{-1}\Sigma_i)+(\mu_i-\mu_j)^{T}\Sigma_j^{-1}(\mu_i-\mu_j))
    \end{aligned}
    \label{computation-cqc}
\end{equation}
Combining Eq.\ref{computation-hqc} and Eq.\ref{computation-cqc}, the $KL$-Divergence between $q(c|x_i,z_i)$ and $q(c|x_j,z_j)$ is
\begin{equation}
    \begin{aligned}
    D_{KL}(q(c|x_i,z_i)||q(c|x_j,z_j))
    =&\int q(c|x_i,z_i) (\log q(c|x_i,z_i)- \log q(c|x_j,z_j))\text{d}c\\
    =&\int \mathcal{N}(c;\mu_i,\Sigma_i) (\log \mathcal{N}(c;\mu_i,\Sigma_i))\text{d}c- \int \mathcal{N}(c;\mu_i,\Sigma_i)(\log \mathcal{N}(c;\mu_j,\Sigma_j))\text{d}c\\
    =&\frac{1}{2}[(\mu_i-\mu_j)^{T}\Sigma_{j}^{-1}(\mu_i-\mu_j)-\log |\Sigma_{j}^{-1}\Sigma_i|+Tr(\Sigma_j^{-1}\Sigma_i)-d]\\
    =&\frac{1}{2}\sum_{l=1}^{d}[(\frac{\mu_{i}^{(l)}-\mu_{j}^{(l)}}{\sigma_j^{(l)}})^{2}-\log(\frac{\sigma_i^{(l)}}{\sigma_j^{(l)}})^{2}+(\frac{\sigma_i^{(l)}}{\sigma_j^{(l)}})^{2}-1]
    \end{aligned}
    \label{DKL-q-q}
\end{equation}
where $l$ denotes the $l$-th element and $\sigma_i^{(l)}$ denotes the positive root of the $l$-th element on the diagonal of covariance matrix $\Sigma_i$.

\subsection{Computation of $D_{KL}(q(c|x,z)||\bar{p}_{k}(c))$}
Let $x_{i}$ and $x_{j}$ denotes the $i$-th and $j$-th sample in dataset $\mathcal{D}_{k}$ with size is $n_k$
\begin{equation}
    \begin{aligned}
    D_{KL}(q(c|x_i,z_i)||\bar{p}_{k}(c))=&\mathbb{E}_{q(c|x_{i},z_{i})}[\log q(c|x_{i},z_{i})-\log\frac{1}{n_k}\sum_{j=1}^{n_{k}}[q(c|x_j,z_j)]\\
    \le&\mathbb{E}_{q(c|x_i,z_i)}[\log q(c|x_i,z_i)-\frac{1}{n_k}\sum_{j=1}^{n_k}\log q(c|x_j,z_j)]\\
    =&\frac{1}{n_k}\sum_{j=1}^{n_k}\mathbb{E}_{q(c|x_i,z_i)}[\log q(c|x_i,z_i)-\log q(c|x_j,z_j)]
    \end{aligned}
\end{equation}
Bringing Eq.\ref{DKL-q-q} we have
\begin{equation}
    \begin{aligned}
    D_{KL}(q(c|x_i,z_i)||\bar{p}_{k}(c))\le\frac{1}{2n_k}\sum_{j=1}^{n_k}\sum_{l=1}^{d}[(\frac{\mu_{i}^{(l)}-\mu_{j}^{(l)}}{\sigma_j^{(l)}})^{2}-\log(\frac{\sigma_i^{(l)}}{\sigma_j^{(l)}})^{2}+(\frac{\sigma_i^{(l)}}{\sigma_j^{(l)}})^{2}-1]
    \end{aligned}
\end{equation}
where $\mu_i$, $\sigma_i$ and $\mu_j$, $\sigma_j$ are outputs of neural networks and they can be differentiated and optimized by gradient based optimization methods.

\section{Experiments}
\subsection{Personalization settings}
\begin{figure}[ht]
    \begin{center}
    \centerline{\includegraphics[width=.5\columnwidth]{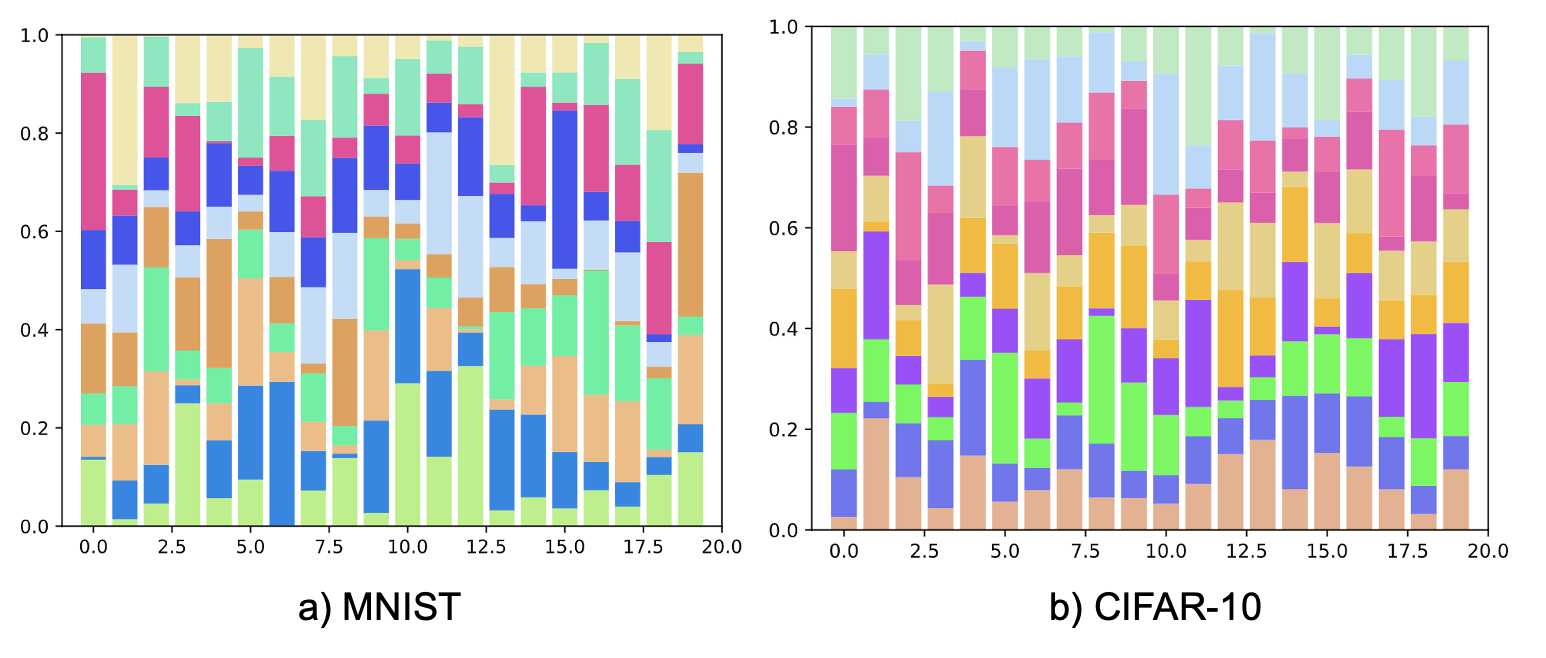}}
    \caption{Class distributions}
    \label{Label distributions}
    \end{center}
\end{figure}

We follow the work in~\cite{DBLP:journals/corr/abs-1909-06335} to allocate samples to 20 clients, with each client having random fractions regarding classes. As described in Fig.\ref{Label distributions}, each column denotes fractions of classes on a client, and each color corresponds to a class. The longer a bar is, the more significant the fraction of the class is on that client.

\subsection{Model architecture}
\begin{figure}[ht]
    \begin{center}
    \centerline{\includegraphics[width=.5\columnwidth]{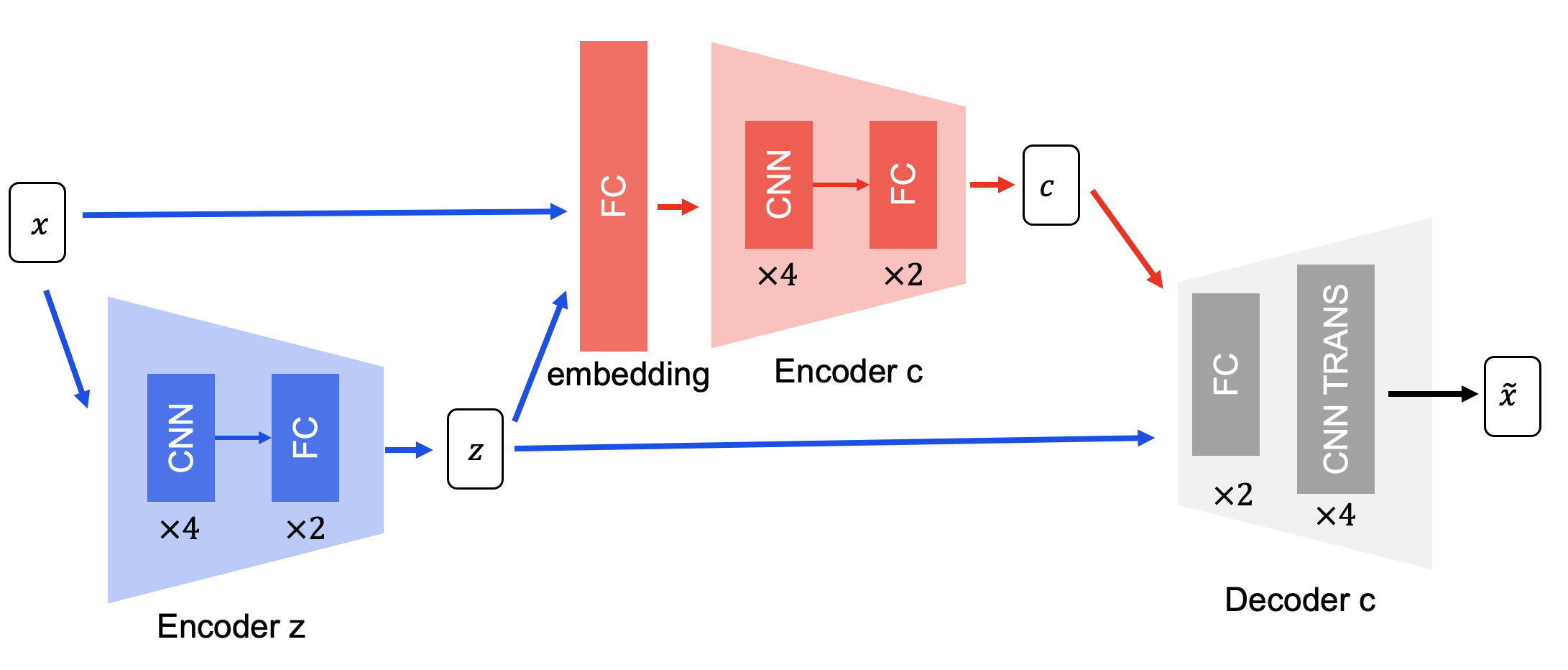}}
    \caption{Model architecture}
    \label{Model architecture}
    \end{center}
\end{figure}

\textbf{FedDVA}: The encoder for representation $z$ consists of a 4-layer CNN backbone and two fully connected embedding layers; The encoder for representation $c$ first combines $z$ and $x$ with an FC layer and then forwards the embedding of $[x;z]$ through a 4-layer CNN backbone and two fully connected embedding layers; The decoder is the reverse of the encoding modules. The model architecture is illustrated in Fig.\ref{Model architecture}, and codes are uploaded along with the supplementary materials.

\textbf{CNN}: To classify samples from MNIST and CIFAR-10, we implemented a CNN consisting of a 4-layer CNN backbone and a 2-FC layer classification head. 

\subsection{Hyperparameters}
On all clients, the batch size is 256 and the learning rate is fixed as 0.001. We trained the global model by 200 communication rounds and 5 epochs during each round. The dimensions of $z$ and $c$ are set to be 4 respectively for reconstruction tasks and 8 for classification tasks. $\xi_{k}$ is set to be 8 times the dimension of $c$. $\alpha$ is 1 and $\beta$ is 0.75.

\subsection{Manifolds of reconstructions}
\begin{figure}[ht]
    \begin{center}
    \centerline{\includegraphics[width=.5\columnwidth]{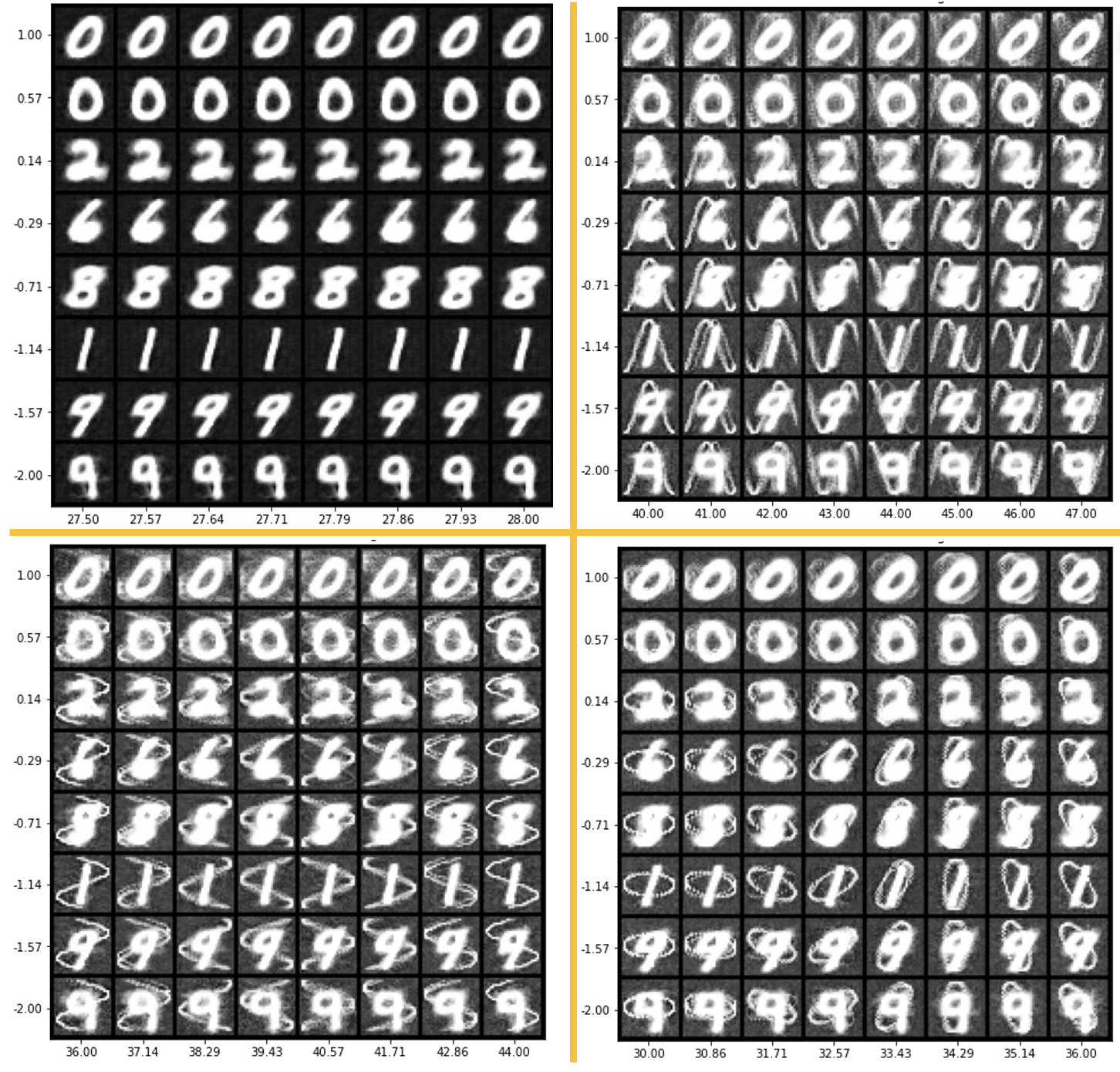}}
    \caption{Reconstructions of MNIST. Each quadrant corresponds to a client. General representations $z$ vary along the vertical axis and personalized representations $c$ vary along the horizontal axis. We can see that $z$ and $c$ are disentangled as digits vary along with changes in $z$ and personalized marks changes with $c$.}
    \label{Reconstructions of MNIST}
    \end{center}
\end{figure}

\begin{figure}[ht]
    \begin{center}
    \centerline{\includegraphics[width=.5\columnwidth]{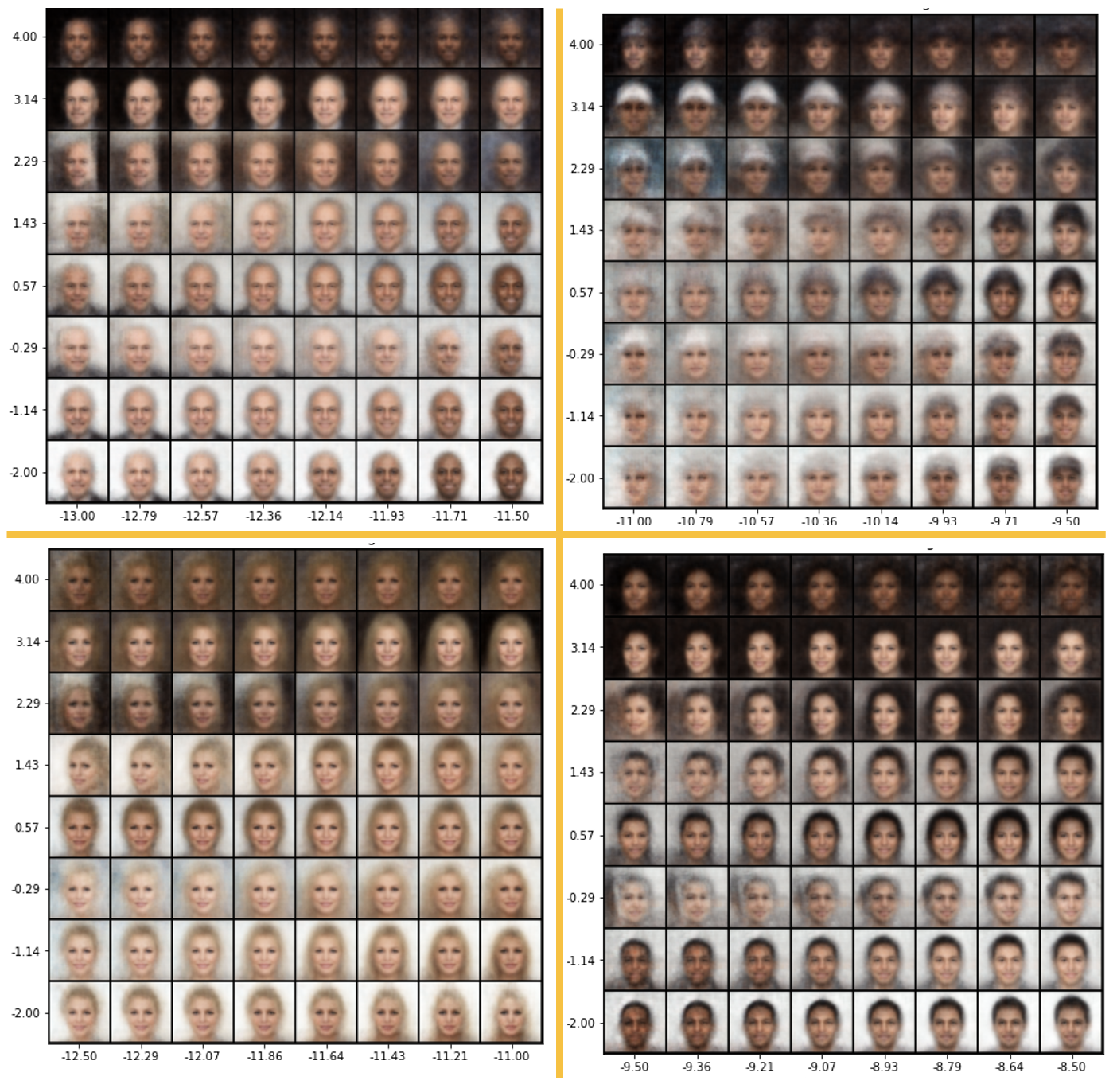}}
    \caption{Reconstructions of CIFAR-10. Each quadrant corresponds to a client. General representations $z$ vary along the vertical axis and personalized representations $c$ vary along the horizontal axis. We can see that $z$ and $c$ are disentangled as faces vary along with changes in $z$ and personalized attributes like hairstyles and skins change with $c$.}
    \label{Reconstructions of CIFAR-10}
    \end{center}
\end{figure}

\label{AppendixExperiments}

\end{document}